\documentclass[letterpaper, 10 pt, conference]{ieeeconf}  
\IEEEoverridecommandlockouts                                                                               
\usepackage{amsmath,amsfonts}
\usepackage{amssymb}
\usepackage{algorithm}
\usepackage[noend]{algorithmic}
\usepackage{color}
\usepackage{array}
\usepackage{textcomp}
\usepackage{stfloats}
\usepackage{verbatim}
\usepackage{graphicx}
\usepackage{cite}
\usepackage[utf8]{inputenc} 
\usepackage[T1]{fontenc}    
\usepackage{hyperref}       
\usepackage{url}            
\usepackage{booktabs}       
\usepackage{amsfonts}       
\usepackage{nicefrac}       
\usepackage{microtype}      
\usepackage{lipsum}
\usepackage{fancyhdr}       
\usepackage{graphicx}       
\usepackage{wrapfig}
\usepackage{booktabs}
\usepackage{multirow}
\usepackage{threeparttable}
\usepackage{graphicx}
\usepackage{multirow}
\usepackage{capt-of}

\title{\LARGE \bf
BARE: Body-Aware Reflexive Evasion for Humanoid Robots
}

\author{Zihao Li, Tong Chen, Hua Song, Zitong Ding, Zihao Xu, Ce Hao, and Weibang Bai$^{\ast}$
\thanks{This work was supported by the Shanghai Pujiang Program under grant 23PJ1408500, Shanghai Kepler Robotics Co., Ltd., and the Key Laboratory of Intelligent Perception and Human-Machine Collaboration (ShanghaiTech University), Ministry of Education.
Corresponding author: Weibang Bai {\itshape (E-mail: wbbai@shanghaitech.edu.cn).}}   
\thanks{Zihao Li, Tong Chen, and Weibang Bai are with the School of Information Science and Technology, ShanghaiTech University, Shanghai 201210, China.}
\thanks{Zihao Xu is with National University of Singapore, Singapore 119077.}
\thanks{Ce Hao is with Zhongguancun Academy, Beijing 100094, China.}
\thanks{Zitong Ding is with Shanghai Jiao Tong University, Shanghai, 200240, China.}
\thanks{Hua Song is with the Shanghai Kepler Robotics Co., Ltd., Shanghai 201210, China.}
}

\begin{document}
\maketitle
\thispagestyle{empty}
\pagestyle{empty}

\begin{abstract}
Dynamic obstacle avoidance is essential for safely deploying humanoid robots in real-world environments, where rapidly approaching objects require immediate evasive responses. 
Recent advances in reflexive evasion have focused primarily on quadrupeds, which typically rely on large-scale whole-body maneuvers. In contrast, humanoids can exploit their rich body articulation to evade through small, localized body adjustments. By moving threatened body regions without repositioning the entire body, these responses may require less execution time and less surrounding free space.
However, local adjustments can alter mass distribution or support conditions, making it challenging to achieve rapid evasion while maintaining whole-body stability.
To this end, we propose BARE (Body-Aware Reflexive Evasion), a humanoid reflexive evasion framework guided by human motion priors that enables rapid local evasion while maintaining whole-body stability. 
A risk-guided motion adapter is designed to select reference motions from a retargeted human evasive motion library based on the threatened body region and to adjust their temporal phases based on collision urgency, allowing the robot to reach evasive postures before potential contact. 
Conditioned on the adapted reference and link-level collision risk, a reflexive evasion policy learns through reinforcement learning to coordinate local body adjustments with whole-body stabilization.
We validate BARE in simulation and real-world experiments, demonstrating its ability to perform targeted local evasion while maintaining whole-body stability.
Videos are available at \url{https://mayberobolab.github.io/BARE-DOA/}.
\end{abstract}

 \begin{figure*}[t]
\centering\includegraphics[width=0.95\textwidth]{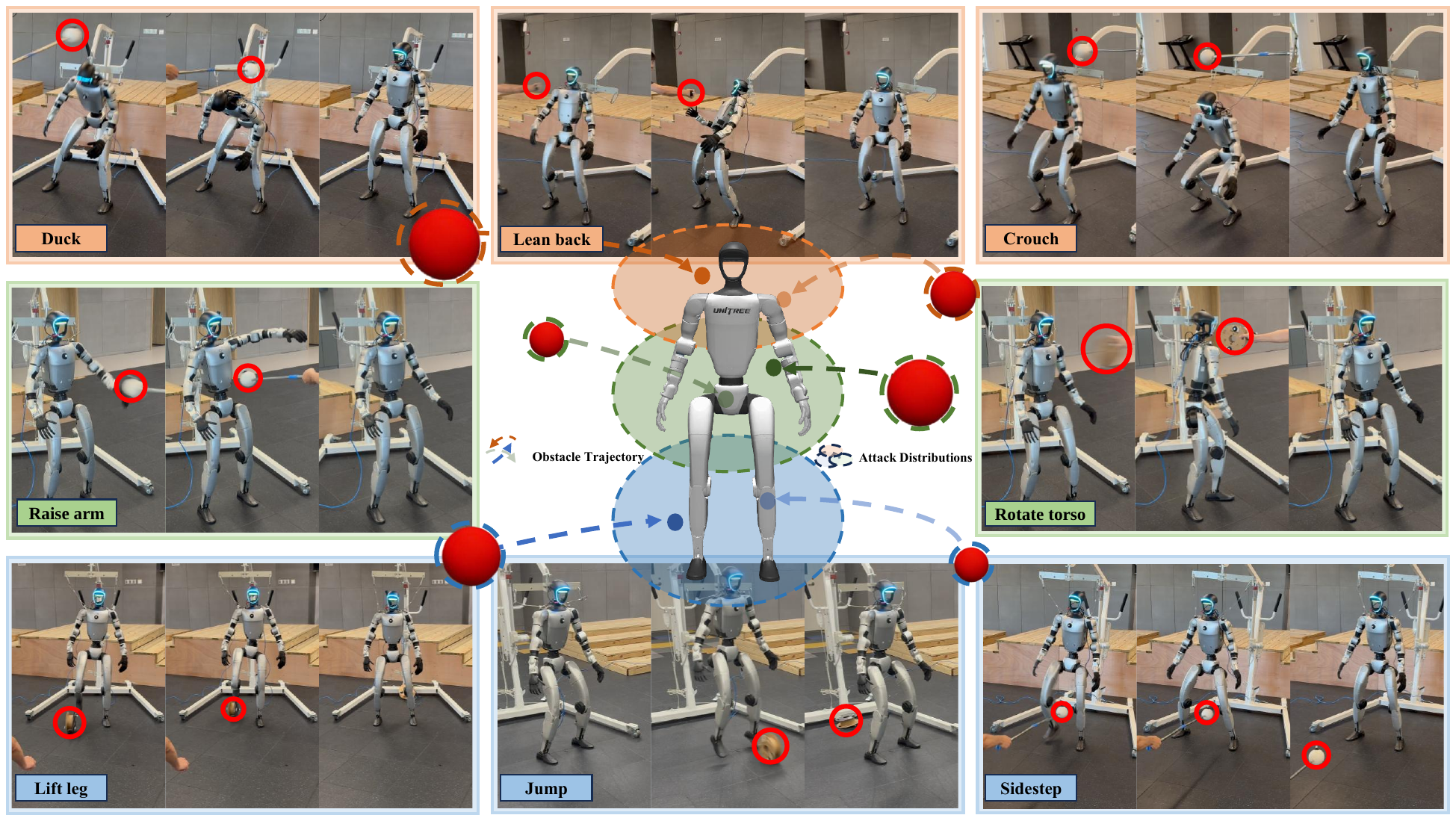}
\caption{Diverse evasive responses of the Unitree G1 humanoid robot using BARE (Body-Aware Reflexive Evasion). The central illustration shows obstacle trajectories toward different body regions. Surrounding image sequences demonstrate ducking, leaning back, crouching, raising an arm, rotating the torso, lifting a leg, jumping, and sidestepping. Ducking lowers the head and inclines the torso, whereas crouching lowers the body through knee flexion. Red circles mark the approaching obstacles.}
\label{fig:teaser}
\vspace{-8pt}
\end{figure*}
\section{Introduction}
\label{sec:intro}
Avoiding collisions with surrounding objects is a fundamental requirement for deploying humanoid robots in real-world environments \cite{sun2025spark,kim2024armor}. Unexpected contact can interrupt task execution and compromise balance, potentially causing falls and damage to the robot or its surroundings\cite{guo2025learning}. Existing approaches address this challenge through visual path planning\cite{lin2025iwalker, wang2024prior}, collision-free footstep generation\cite{gaspard2024footstepnet}, and online gait adjustment\cite{narkhede2023overtaking,li2026s}, enabling humanoid robots to navigate around clutter and moving obstacles.

However, walking-based avoidance may be insufficient when dynamic obstacles approach rapidly at close range. Planning delays and the time required to execute movements within actuator limits can leave the robot unable to avoid contact in time. In such situations, robots need to evade obstacles reflexively by rapidly adjusting their bodies to create clearance.
Recent work has advanced this capability on quadrupeds. REBot learns evasion and recovery policies for rapid obstacle avoidance, while APREBot and UEREBot extend these capabilities through active perception and navigation--evasion coordination in complex environments \cite{xu2025rebot,xu2025aprebot,xu2026uerebot}.
However, these methods are not fully suited to humanoid robots. With relatively few degrees of freedom, quadrupedal robots are typically guided by existing policies to evade obstacles through large-scale whole-body maneuvers, such as jumping, lateral stepping, or retreating.
In contrast, humanoid robots have more degrees of freedom and can exploit local body adjustments, such as retracting an arm or lifting a leg, to create clearance without repositioning the entire body. Such small-scale, targeted responses can reduce both the execution time and the free space required for evasion, which is especially advantageous when obstacles approach rapidly at close range.
Nevertheless, local evasive motions can alter the robot's mass distribution or support configuration, making rapid local evasion while maintaining whole-body stability a key challenge.

To address this challenge, we propose \textbf{BARE} (Body-Aware Reflexive Evasion), a framework for humanoid reflexive evasion guided by human motion priors. It adjusts coordinated retargeted human evasive motions based on collision risk across body regions, enabling targeted local responses within a short response window while maintaining whole-body stability. 
The framework comprises two components. 
A Risk-Guided Motion Adapter evaluates link-level collision risk from the robot state and relative obstacle motion. When risk exceeds a threshold, the adapter selects a reference from a library of retargeted human evasive motions according to the threatened body region and adjusts its phase according to collision urgency so that key evasive poses are reached in time. 
A Reflexive Evasion Policy generates joint-position targets from the adapted reference, link-level collision risk, and robot and obstacle states. Through reference-guided reinforcement learning, the policy learns to move threatened body regions out of the obstacle's path while maintaining whole-body stability.

We evaluate BARE in simulation and on a physical Unitree G1 to assess evasion performance within a short response window across obstacle arrival times, approach directions, and target heights. Ablation studies evaluate the contributions of risk-conditioned motion selection, phase adaptation, and the reward design.

The main contributions are as follows:
\begin{itemize}
\item We formulate humanoid dynamic obstacle avoidance as a risk-conditioned reflexive control problem, emphasizing local evasion of threatened body regions and whole-body stability.
\item We propose BARE, a framework comprising a Risk-Guided Motion Adapter and a Reflexive Evasion Policy, which uses human motion priors to guide the learning of targeted evasive responses according to link-level collision risk.
\item We conduct simulation and real-world experiments on the Unitree G1 humanoid to evaluate BARE's effectiveness in rapid local evasion.
\end{itemize}

 \section{Related Work}
Human motion tracking and learned motion priors provide structured guidance for coordinated skills in physically simulated characters, as demonstrated by DeepMimic and AMP \cite{peng2018deepmimic,peng2021amp}. H2O and OmniH2O extend human-motion learning to real humanoid whole-body teleoperation, while ASAP addresses dynamics mismatch between simulation and real robots for agile skills \cite{he2024learning,he2024omnih2o,he2025asap}. BeyondMimic further develops versatile control through motion tracking and guided diffusion \cite{liao2026beyondmimic}; parkour and table-tennis systems demonstrate terrain traversal and rapid whole-body interaction \cite{zhuang2024humanoid,su2025hitter}.
Alongside these advances in humanoid motion capabilities, progress has also been made in safe navigation. iWalker learns visual path and footstep planning, FootstepNet generates collision-free footstep sequences, and model predictive contouring control enables Digit to overtake moving obstacles \cite{lin2025iwalker,gaspard2024footstepnet,narkhede2023overtaking}. These approaches avoid obstacles by adjusting walking paths and gait. Even with rapid planning updates, executing the required motion may exceed the remaining response window against close-range, rapidly approaching threats.

To handle rapidly approaching dynamic obstacles, researchers have explored avoidance methods that connect environmental perception with immediate motion responses. For UAVs, event-camera-based perception and end-to-end learning policies have demonstrated rapid responses to dynamic obstacles \cite{falanga2020dynamic,wang2026keypoint,fan2025flying}. For legged robots, REBot combines reinforcement learning-based evasion and recovery policies, enabling quadrupeds to avoid incoming obstacles within a short response window and regain stability \cite{xu2025rebot}. APREBot introduces active perception combining LiDAR and cameras to improve awareness of dynamic threats from different directions \cite{xu2025aprebot}. UEREBot further coordinates navigation and reflexive evasion with safety constraints, supporting both goal progress and high-speed obstacle avoidance in unstructured environments \cite{xu2026uerebot}. These quadrupedal methods predominantly evade threats through whole-body maneuvers such as jumping, lateral displacement, or retreating. The richer articulation of humanoids offers opportunities for localized responses, but these motions must remain coordinated with whole-body balance. Exploiting this capability for timely, stable, and targeted evasion requires further investigation.
 
\begin{figure}[t]
\centering
\includegraphics[width=\columnwidth]{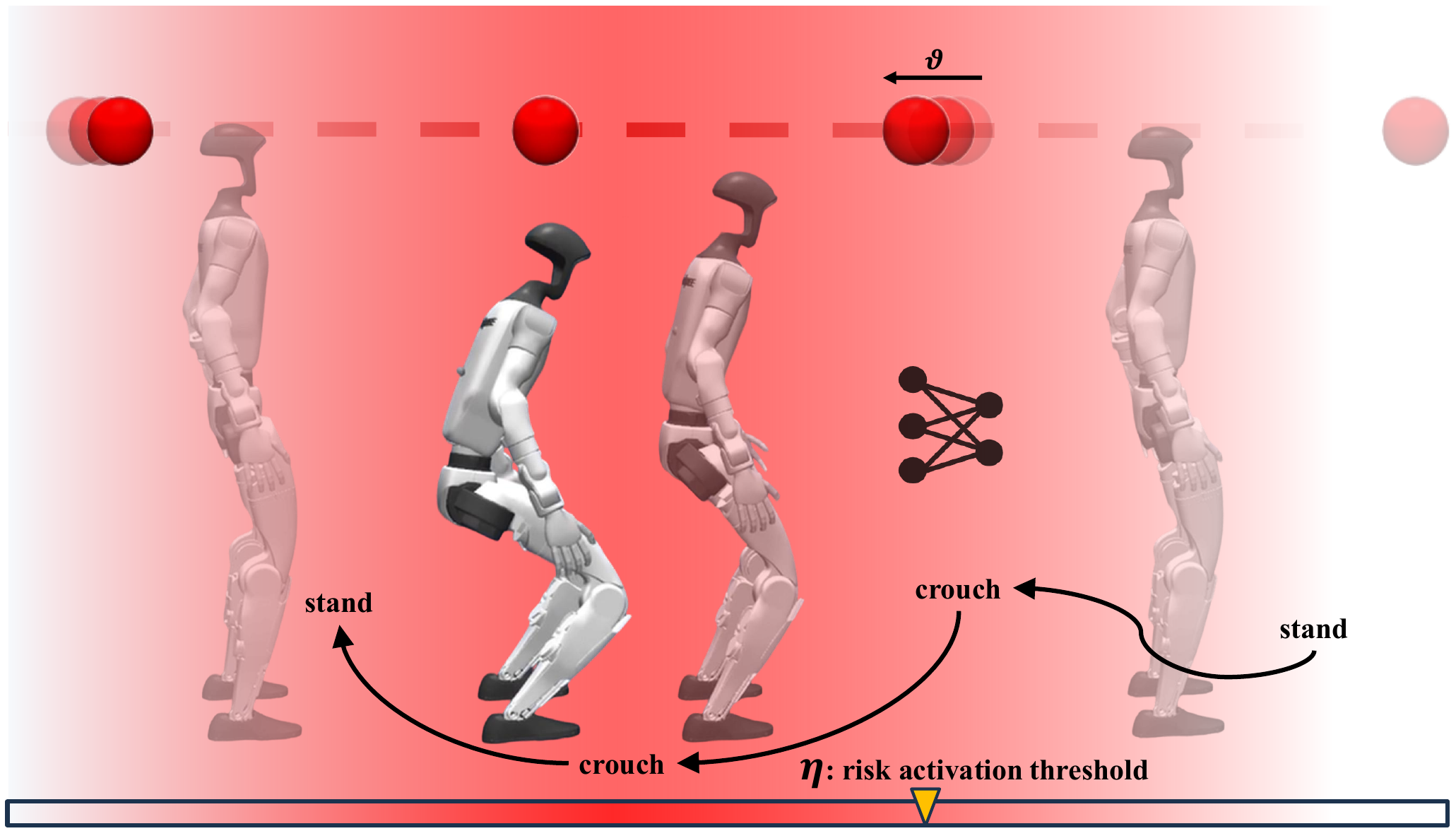}
\caption{\textbf{Task overview.} As an obstacle approaches, the humanoid transitions from standing to an evasive posture and subsequently returns to standing. The red shading illustrates collision risk, and the risk activation threshold marks the onset of the evasion response. The sequence follows the obstacle motion from right to left.}
\label{fig:task}
\vspace{-8pt}
\end{figure}
\begin{figure*}[t]
\centering
\includegraphics[width=\textwidth]{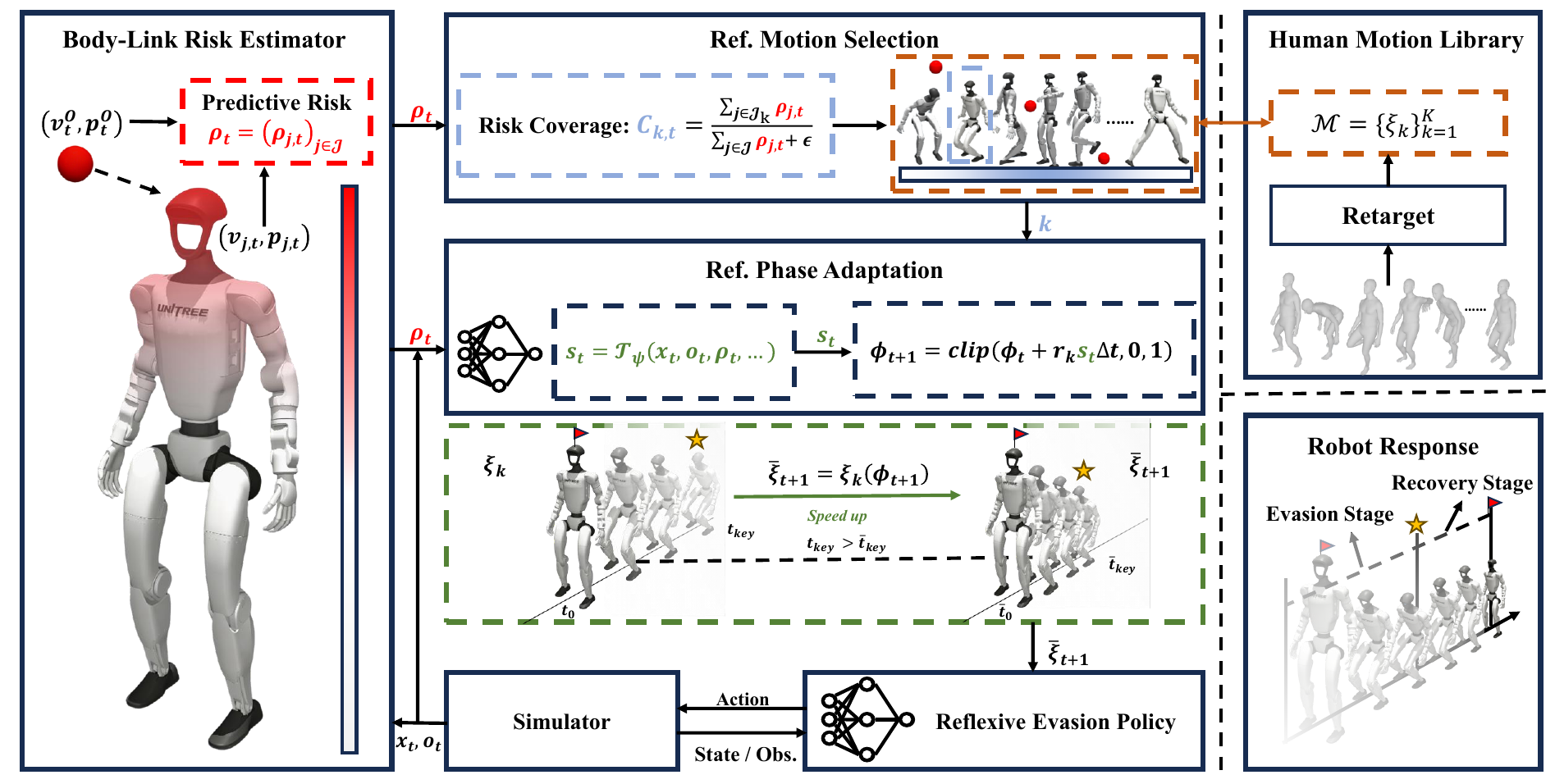}
\caption{\textbf{Overview of BARE (Body-Aware Reflexive Evasion).} BARE comprises a risk-guided motion adapter and a reflexive evasion policy. The adapter consists of the Body-Link Risk Estimator, Ref. Motion Selection, and Ref. Phase Adaptation modules shown in the figure. Human motions are retargeted to form a reference library. Based on link-level collision risks, the adapter selects a reference motion and adjusts its phase. The adapted reference guides the policy in executing evasion and recovery. The phase-adaptation illustration shows an example of accelerating the reference motion, while the lower loop depicts policy interaction with the simulator.}
\label{fig:method}
\vspace{-8pt}
\end{figure*}
\section{Problem Formulation}
\label{sec:problem}
As illustrated in Fig.~\ref{fig:task}, we study humanoid evasion of a rapidly approaching dynamic obstacle within a short response window. The robot must generate coordinated responses to threats affecting different body regions, avoid collision while preserving balance, and recover to a stable state after the threat passes. Consider a humanoid with $n$ actuated joints and state $x_t=(p_t^R,R_t^R,v_t^R,\omega_t^R,q_t,\dot q_t)$, comprising the base position, orientation, linear and angular velocities, and joint positions and velocities. A single dynamic obstacle is approximated as a sphere with state $o_t=(p_t^O,v_t^O,r^O)$, specifying its position, velocity, and effective radius. Positions and velocities are expressed in the world frame unless otherwise stated. At each control step, the controller uses robot proprioception and obstacle observations to generate joint-position targets $u_t\in\mathbb R^n$, which are executed by low-level PD controllers. We consider dynamic threats approaching different body regions from varying directions and heights.

Let $\mathcal J$ denote all body links included in obstacle collision detection, $\mathcal B_j(x_t)$ the space occupied by link $j$, and $\mathcal O_t=\{p\in\mathbb R^3:\|p-p_t^O\|\le r^O\}$ the obstacle region. The signed surface clearance is $d_j(t)=\operatorname{sdist}(\mathcal B_j(x_t),\mathcal O_t)$, and the minimum whole-body clearance is $d_{\min}(t)=\min_{j\in\mathcal J}d_j(t)$. Positive, zero, and negative values indicate separation, contact, and penetration, respectively. Collision-free whole-body motion therefore requires positive clearance for every relevant link, rather than only for the initially threatened region.

Local evasion creates clearance primarily through motion of the threatened body region and its associated joints, with whole-body coordination maintaining balance; larger-scale body motion is permitted when necessary. Success requires $d_{\min}(t)>0$ throughout the obstacle's approach and passage, no falls or instability terminations during evasion and recovery, and a return to stability after the threat passes.
 \section{Method}
\label{sec:method}
This section presents BARE, a framework for humanoid reflexive evasion. We first outline the framework and its operation (Sec.~\ref{sec:overview}), then introduce the risk-guided motion adapter, which selects reference motions and adjusts their timing according to link-level collision risk (Sec.~\ref{sec:adapter}). Finally, we describe how the reflexive evasion policy learns local evasion and whole-body coordination from the adapted references, together with its reward design and training procedure (Sec.~\ref{sec:policy}).

\subsection{Framework Overview}
\label{sec:overview}

BARE comprises a risk-guided motion adapter and a reflexive evasion policy (Fig.~\ref{fig:method}). The adapter integrates the Body-Link Risk Estimator, Ref. Motion Selection, and Ref. Phase Adaptation modules shown in the figure. Offline, human evasive motions are recovered from videos and retargeted to build a library containing local evasive motions and whole-body displacement maneuvers. Online, the adapter estimates link-level collision risks from relative robot--obstacle motion. Once risk reaches the activation threshold, it selects a reference suited to the threat and adapts its phase to the obstacle motion. Conditioned on the adapted reference, the link-level collision-risk vector, and robot and obstacle states, the reflexive evasion policy produces joint-position targets to execute evasion and recovery.

\subsection{Risk-Guided Motion Adapter}
\label{sec:adapter}

\textbf{Human motion references.} We recover human motions from evasion videos using GVHMR~\cite{shen2024gvhmr} and retarget them to the robot using GMR~\cite{araujo2025retargeting}, yielding a motion library $\mathcal M=\{\xi_k\}_{k=1}^{K}$. Each reference trajectory $\xi_k$ is annotated with a motion type $\mathrm{type}_k\in\{\mathrm{local},\mathrm{displacement}\}$ and a set of addressed body links $\mathcal J_k$. The two types denote local evasive motions and whole-body displacement maneuvers, respectively. Each motion $\xi_k(\phi)$ is indexed by normalized phase $\phi\in[0,1]$ and contains joint and base reference states, with annotations for motion onset, a key evasive pose, and the recovery endpoint. The library includes local evasive motions, such as raising an arm or lifting a leg, as well as whole-body displacement maneuvers, such as lateral movement and retreat. Local evasive motions primarily reposition the threatened body region, whereas whole-body displacement maneuvers move the robot base to create clearance. Both types retain whole-body coordination for balance.

\textbf{Link-level collision risk.} To determine where a response is needed and how urgently it must occur, we predict short-term motion for individual body links. Online risk estimation uses spherical approximations of link geometry. Let $p_{j,t}$, $v_{j,t}$, and $r_j$ denote the center, velocity, and radius of the approximating sphere of link $j$. Define relative position $a_{j,t}=p_t^O-p_{j,t}$ and relative velocity $b_{j,t}=v_t^O-v_{j,t}$. Under a short-term constant-velocity approximation, the time to closest approach within horizon $H$ and the corresponding surface clearance are:
\begin{equation}
\begin{aligned}
\tau_{j,t}^{*}&=\operatorname{clip}\!\left(-\frac{a_{j,t}^{\top}b_{j,t}}{\|b_{j,t}\|^2},0,H\right),\\
\hat d_{j,t}&=\|a_{j,t}+\tau_{j,t}^{*}b_{j,t}\|-(r_j+r^O).
\end{aligned}
\end{equation}

We set $\tau_{j,t}^{*}=0$ when relative speed is near zero. The current clearance between the approximating spheres is $\tilde d_{j,t}=\|a_{j,t}\|-(r_j+r^O)$, and the closing speed is $c_{j,t}=-a_{j,t}^{\top}b_{j,t}/\max(\|a_{j,t}\|,\epsilon)$. We combine immediate proximity, predicted clearance, and temporal urgency as:
\begin{equation}
\begin{aligned}
I_{j,t}&=\sigma\!\left(\frac{m-\tilde d_{j,t}}{s_d}\right),\\
P_{j,t}&=\sigma\!\left(\frac{m-\hat d_{j,t}}{s_d}\right)
\sigma\!\left(\frac{c_{j,t}}{s_v}\right)e^{-\tau_{j,t}^{*}/s_\tau},\\
\rho_{j,t}&=1-(1-I_{j,t})(1-P_{j,t}),
\end{aligned}
\end{equation}
where $\sigma$ is the sigmoid function, $m$ is a safety margin, and $s_d,s_v,s_\tau$ set the respective scales. Each body region comprises one or more links, and link-level collision risks represent the distribution of threats across these regions. The risk vector $\boldsymbol\rho_t=(\rho_{j,t})_{j\in\mathcal J}$ preserves the spatial distribution of threats across the body, while $\rho_t^{\max}=\max_j\rho_{j,t}$ summarizes whole-body threat severity and is supplied to the adapter alongside the risk vector. Clearances between the approximating spheres support online risk estimation and do not replace the whole-body collision criterion in the problem formulation.

\textbf{Reference motion selection.} When the maximum link-level collision risk $\rho_t^{\max}$ reaches threshold $\eta$, the adapter uses a rule-based mechanism to select a reference according to the risk distribution and obstacle approach direction. Let $\mathcal J_k$ denote the set of body links addressed by motion $k$. The risk coverage of motion $k$ is defined as:
\begin{equation}
C_{k,t}=\frac{\sum_{j\in\mathcal J_k}\rho_{j,t}}{\sum_{j\in\mathcal J}\rho_{j,t}+\epsilon}.
\end{equation}

If the maximum coverage among local motions reaches threshold $\gamma$, local motions addressing high-risk links become candidates; otherwise, whole-body displacement maneuvers such as lateral movement and retreat are considered. A high-risk link has a risk value no lower than a predefined fraction of the peak risk across the body. The adapter then uses the approach direction to determine a reference from the candidates, matching the response to the threatened region and obstacle motion. The selected reference remains fixed throughout an evasion and recovery response. Before activation, the robot maintains its nominal standing posture.

\textbf{Reference phase adaptation.} A learned policy $\mathcal T_\psi$ adapts reference phase. Its input $z_t^{\mathrm{ad}}$ contains proprioception, relative obstacle state, link-level collision risks, the selected reference, and current phase. The policy predicts a phase-rate scaling factor $s_t$; the scheduler combines it with the motion-specific nominal phase rate $r_k$ and control interval $\Delta t$ to update the reference:
\begin{equation}
\begin{aligned}
s_t&=\mathcal T_\psi(z_t^{\mathrm{ad}}),\\
\phi_{t+1}&=\operatorname{clip}(\phi_t+r_k s_t\Delta t,0,1),\\
\bar\xi_{t+1}&=\xi_k(\phi_{t+1}).
\end{aligned}
\end{equation}

With $s_t>0$, the scheduler advances the phase in the reference's temporal order until the endpoint. The phase policy is optimized through reinforcement learning for executed evasion, recovery, and execution cost, while the selection rule remains fixed.

Phase adaptation aligns the effective evasion phase of the reference with obstacle passage while accounting for stability during recovery.

\begin{figure*}[t]
\centering
\includegraphics[width=\textwidth]{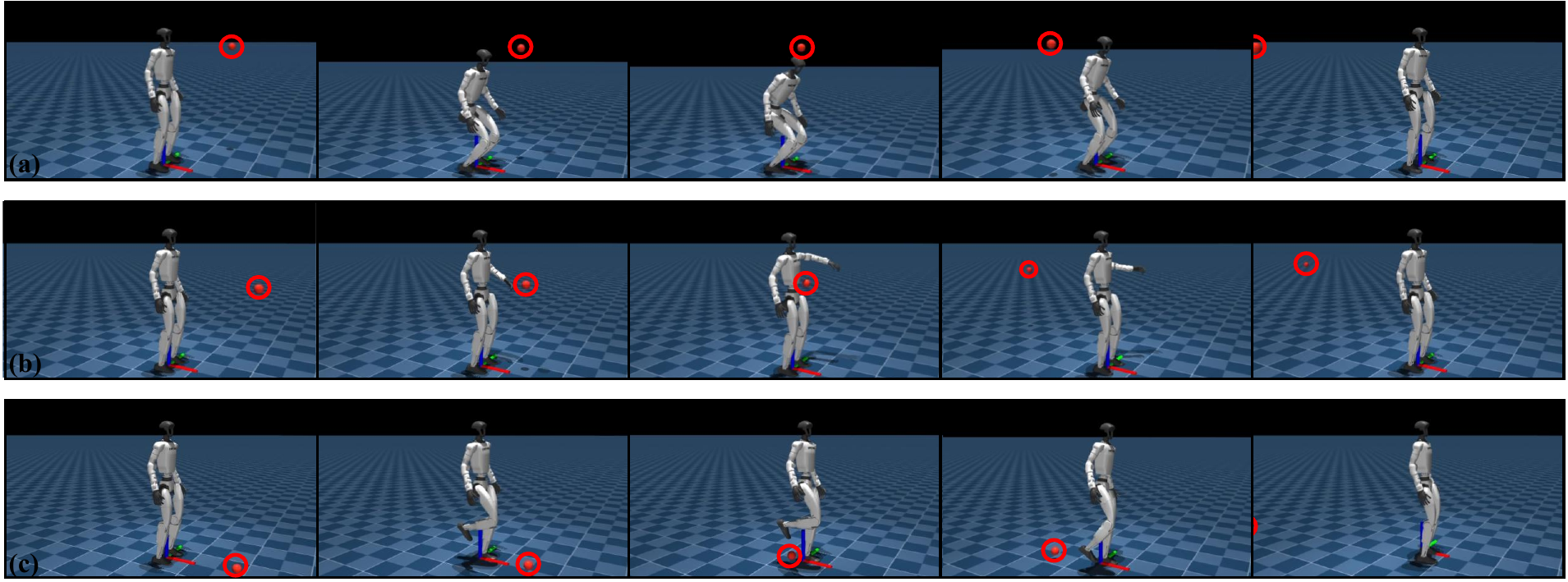}
\caption{Representative simulated evasion sequences of BARE against projectiles
targeting different body heights. From top to bottom, the rows show
attacks targeting the upper, middle, and lower body regions. Each
sequence progresses from left to right through evasion and recovery.}
\label{fig:sim-demo}
\vspace{-8pt}
\end{figure*}

\subsection{Reflexive Evasion Policy}
\label{sec:policy}

The adapted motion provides a coordinated reference, but successful evasion also depends on robot dynamics, execution errors, and immediate obstacle motion. The reflexive evasion policy conditions on proprioception, relative obstacle state, link-level collision risk, and reference state to convert the reference into executable evasive behavior. Denoting its input by $z_t$, following residual learning that combines a nominal signal with a learned correction~\cite{johannink2019residual}, the policy produces a joint residual $\delta q_t=\pi_\theta(z_t)$ and forms the control target around the reference joint position $\bar q_t$ as:
\begin{equation}
u_t=\bar q_t+\beta\,w_t\odot\delta q_t,
\end{equation}

where $\beta$ is the residual scale, $w_t$ is a joint-wise binary residual mask associated with the selected reference motion, and $\odot$ denotes element-wise multiplication. This mask concentrates corrections on joints relevant to the current response, while the reference itself retains motion needed for whole-body coordination. Local evasion therefore does not require the remaining body regions to stay still: lifting a leg, for example, may involve coordinated adjustments of the support leg and torso.

The residual mask is determined by the selected reference motion. Let $A_{k,i}(\phi_t)$ denote the activity weight of joint $i$ in reference motion $k$. During evasion and recovery, we set $w_{i,t}=\mathbf 1[A_{k,i}(\phi_t)>0]$, and disable residual corrections upon completion. The mask restricts residual corrections, while the reference motion retains whole-body coordination for balance.

Policy learning considers collision avoidance, reference execution, and motion stability. To clarify their roles, we group the reward terms by function:
\begin{equation}
\begin{aligned}
r_t= & \lambda_e r_t^{\mathrm{evasion}}+\lambda_r r_t^{\mathrm{reference}} \\
&  + \lambda_l r_t^{\mathrm{locality}}+\lambda_{reg}r_t^{\mathrm{regularization}},
\end{aligned}
\end{equation}
where $\lambda_e,\lambda_r,\lambda_l,\lambda_{reg}$ are the weights of the corresponding reward groups.

The evasion terms penalize insufficient clearance at threatened links and actual robot--obstacle contact, encouraging timely clearance creation. Contact detection covers the entire body so that moving the initially threatened region does not hide a collision involving another link. Reference terms guide position and velocity tracking for relevant joints, organizing high-dimensional exploration around coordinated human evasive motions \cite{liao2026beyondmimic}. For lateral movement and retreat, reference base displacement is also tracked to preserve the motion required for whole-body displacement maneuvers.

The locality term accounts for the number of joints actually moved, horizontal base displacement, and joint power. It is enabled when the collision-free clearance condition is satisfied, encouraging local and efficient execution among safe responses. Regularization terms penalize action changes, joint-limit violations, and self-collisions, with additional penalties for falls or instability terminations. Together, these objectives encourage threatened-region evasion while retaining whole-body balance and recovery. Reward trade-offs do not provide formal collision-avoidance or stability guarantees; both are evaluated using the criteria in the problem formulation.

We optimize the learned policies with proximal policy optimization (PPO)~\cite{schulman2017proximal}. Training combines staged learning with joint optimization to coordinate reference execution and risk-conditioned motion adaptation. Staged learning establishes the capabilities required by each component, while joint optimization further coordinates phase adaptation and executed evasion, with reference selection rules held fixed.

\section{Simulation Experiments}

We organize our simulation experiments around three questions: \textbf{Q1:} Can BARE improve humanoid evasion success under limited response time? (Sec.~\ref{sec:main}) \textbf{Q2:} Can BARE achieve successful evasion with smaller base displacement and lower peak mechanical power than the baselines? (Sec.~\ref{sec:main}) \textbf{Q3:} How do risk-conditioned motion selection, phase adaptation, and the locality reward contribute to evasion performance? (Sec.~\ref{sec:ablation})

\subsection{Experimental Setup}

\textbf{Task.} We use a Unitree G1 humanoid with 29 actuated joints in mjlab with the MuJoCo backend \cite{todorov2012mujoco}. Each trial starts from the same nominal standing posture. Threatening projectiles are randomly generated within predefined body-target regions, with horizontal bearings spanning $[0^\circ,360^\circ)$. Projectile size is held consistent across methods, and initial positions and velocities are constructed consistently from target positions, directions, and arrival times. Arrival time $T_{\mathrm{arr}}$ denotes the interval from attack onset to predicted first contact if the robot maintains its nominal standing posture. We use three groups: short $[0.10,0.50)$ s, medium $[0.50,1.50)$ s, and long $[1.50,3.00]$ s. Trials are randomly sampled within each interval with seeds independent of training. We retain attacks predicted to hit the nominal posture. Main comparisons and ablations share the attack distribution; all methods use the same robot model, attacks, obstacle-state information, and success criteria.

\begin{figure*}[t]
\centering
\includegraphics[width=0.9\textwidth]{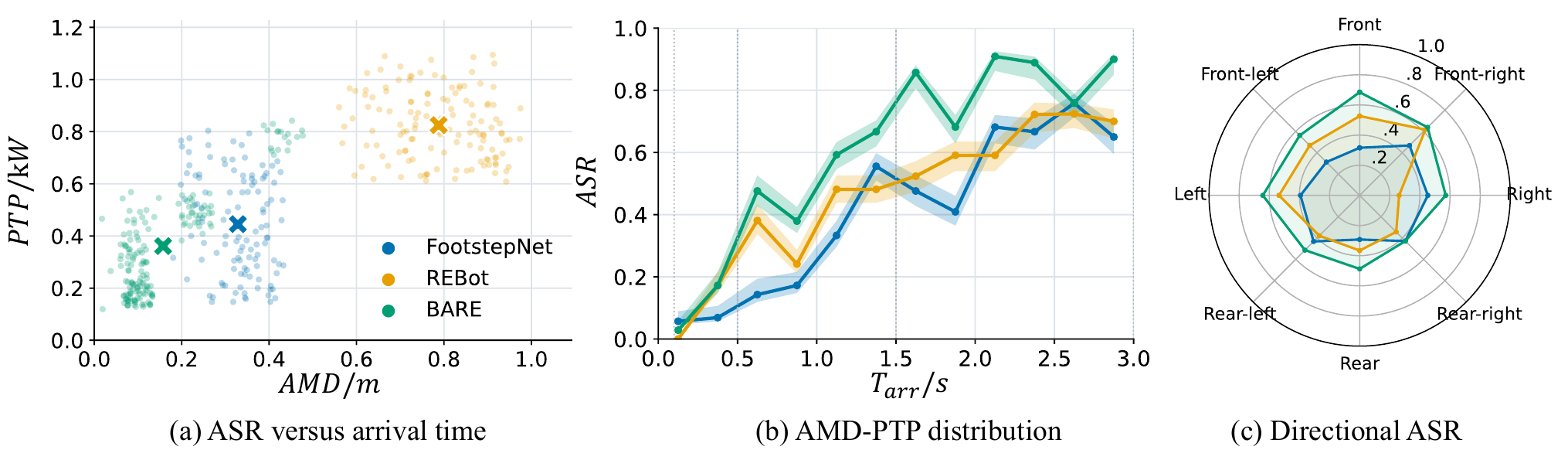}
\vspace{-2mm}
\caption{Main comparison of FootstepNet, REBot, and BARE. (a) AMD and PTP of successful trials; crosses denote method means. Successful subsets differ across methods, so costs should be interpreted alongside ASR. (b) ASR in 0.25-s arrival-time bins. (c) ASR pooled over arrival times in eight horizontal sectors spanning $\pm22.5^\circ$ around each indicated bearing.}
\label{fig:main-results}
\vspace{-2mm}
\end{figure*}

\textbf{Metrics.} We report avoidance success rate (ASR), peak total power (PTP), and average movement distance (AMD). ASR is defined as $\mathrm{ASR}=N_{\mathrm{success}}/N_{\mathrm{total}}$, where the numerator counts successful trials and the denominator counts all trials. Success requires collision-free obstacle passage, no instability during evasion and recovery. ASR is reported as a proportion from 0 to 1; all tabulated values use two decimal places. PTP is $\mathrm{PTP}=\max_t\sum_i|\tau_i(t)\dot q_i(t)|$. Joint mechanical power is evaluated at each 5-ms physics step, and PTP is the maximum summed absolute joint power over the evaluation interval, reported in kW. AMD measures the mean horizontal base displacement from the activation position during evasion and recovery, in metres. ASR includes all valid attacks; PTP and AMD are averaged over successful trials.

\begin{table}[!t]
\centering
\renewcommand{\arraystretch}{1.1}
\caption{Simulation Experiment Results}
\label{tab:1}
{\scriptsize
\resizebox{\columnwidth}{!}{
\begin{threeparttable}
\begin{tabular}{llccc}
\toprule
$T_{\mathrm{arr}}$ / s & Metric & FootstepNet \cite{gaspard2024footstepnet} & REBot \cite{xu2025rebot} & BARE \\
\midrule
\multirow{3}{*}{$[0.10,0.50)$}
 & ASR$\uparrow$ & 0.08 & 0.10 & \textbf{0.12} \\
 & PTP$\downarrow$ & 0.54 & 0.86 & \textbf{0.41} \\
 & AMD$\downarrow$ & 0.23 & 0.89 & \textbf{0.14} \\
\midrule
\multirow{3}{*}{$[0.50,1.50)$}
 & ASR$\uparrow$ & 0.31 & 0.39 & \textbf{0.53} \\
 & PTP$\downarrow$ & 0.52 & 0.84 & \textbf{0.42} \\
 & AMD$\downarrow$ & 0.29 & 0.75 & \textbf{0.15} \\
\midrule
\multirow{3}{*}{$[1.50,3.00]$}
 & ASR$\uparrow$ & 0.61 & 0.64 & \textbf{0.83} \\
 & PTP$\downarrow$ & 0.41 & 0.82 & \textbf{0.33} \\
 & AMD$\downarrow$ & 0.35 & 0.80 & \textbf{0.16} \\
\bottomrule
\end{tabular}
\begin{tablenotes}[flushleft]
\scriptsize
\item[] ASR: avoidance success rate;
PTP: peak total power (kW);
AMD: average movement distance (m).
PTP and AMD are averaged over successful trials.
Best values are bold.
\end{tablenotes}
\end{threeparttable}
}}
\vspace{-5mm}
\end{table}

\textbf{Baselines.} REBot (G1, adapted) follows the reinforcement-learning evasion and recovery approach of REBot, using a reference-free G1 policy trained on the common attack distribution \cite{xu2025rebot}. FootstepNet (G1) serves as a navigation-based avoidance baseline based on footstep planning; its original method uses reinforcement learning to generate feasible collision-free footstep sequences \cite{gaspard2024footstepnet}. Each method retains its own response decision mechanism.

\subsection{Main Results}\label{sec:main}

Figure \ref{fig:sim-demo} shows representative responses. Figure \ref{fig:main-results}(b) plots ASR in 0.25-s arrival-time bins, while Table~\ref{tab:1} reports the three predefined groups. BARE obtains ASR values of 0.12, 0.53, and 0.83 in the short, medium, and long groups, respectively, compared with 0.08, 0.31, and 0.61 for FootstepNet and 0.10, 0.39, and 0.64 for REBot. Although all methods are challenged by the shortest arrivals, BARE is best in all three predefined arrival-time groups and improves consistently as arrival time increases.

Among successful trials, BARE has PTP values of 0.41, 0.42, and 0.33 kW and AMD values of 0.14, 0.15, and 0.16 m across the three groups, lower than both baselines in every group. Figure \ref{fig:main-results}(a) shows that BARE samples cluster in the low-power, low-displacement region, while Fig. \ref{fig:main-results}(c) shows that its advantage spans multiple horizontal approach sectors. BARE therefore improves evasion success while achieving smaller base displacement and lower peak mechanical power among successful trials.

\subsection{BARE Ablations}\label{sec:ablation}

We separately ablate risk-conditioned motion selection, phase adaptation, and locality reward. At risk activation, w/o motion selection samples uniformly from the full reference library, removing risk-coverage and approach-direction filtering while retaining learned phase adaptation. The w/o phase adaptation variant retains risk-conditioned motion selection but fixes phase progression to the nominal rate. These two variants replace decisions at test time using the same BARE checkpoint; w/o locality reward removes the locality reward and is retrained. The remaining risk trigger, observations, and control interfaces are unchanged.

\begin{figure*}[t]
\centering
\includegraphics[width=\textwidth]{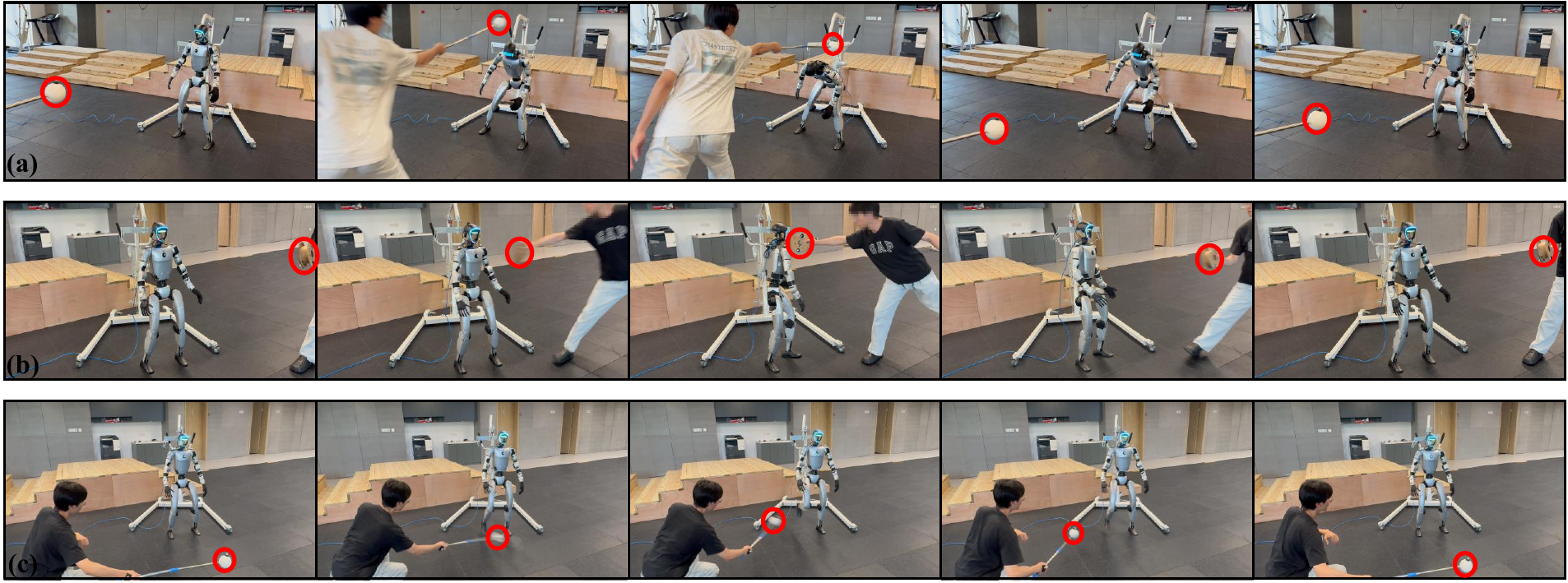}
\caption{Representative real-robot evasion sequences of BARE on Unitree G1. Rows show upper-, middle-, and lower-body attacks from top to bottom, with time progressing from left to right.}
\label{fig:real-demo}
\end{figure*}

\begin{figure}[t]
\centering
\includegraphics[width=\columnwidth]{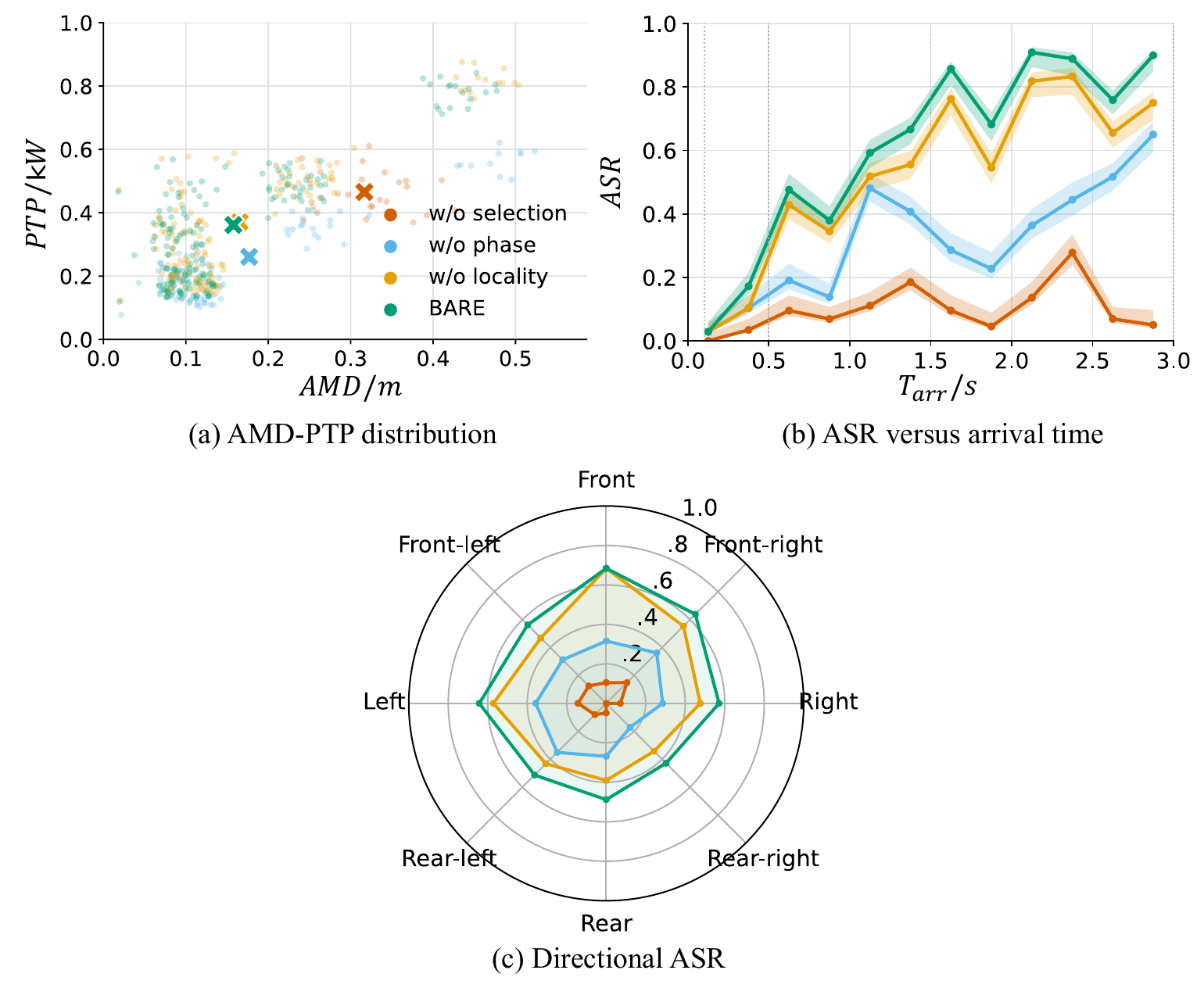}
\caption{Comparison of BARE and its three ablations. (a) AMD and PTP of successful trials; crosses denote method means. Successful subsets differ across methods, so costs should be interpreted alongside ASR. (b) ASR in 0.25-s arrival-time bins. (c) ASR pooled over arrival times in eight horizontal sectors spanning $\pm22.5^\circ$ around each indicated bearing.}
\label{fig:ablation-results}
\vspace{-8pt}
\end{figure}

\begin{table}[t]
\centering
\renewcommand{\arraystretch}{1.1}
\caption{Ablation Experiment Results}
\label{tab:3}
{\scriptsize
\resizebox{\columnwidth}{!}{
\begin{threeparttable}
\begin{tabular}{llcccc}
\toprule
$T_{\mathrm{arr}}$ / s & Metric
& w/o selection & w/o phase & w/o locality & BARE \\
\midrule
\multirow{3}{*}{$[0.10,0.50)$}
& ASR $\uparrow$ & 0.02 & 0.08 & 0.08 & \textbf{0.12} \\
& PTP (kW) $\downarrow$ & 0.45 & \textbf{0.23} & 0.42 & 0.41 \\
& AMD (m) $\downarrow$ & 0.26 & 0.16 & 0.15 & \textbf{0.14} \\
\midrule
\multirow{3}{*}{$[0.50,1.50)$}
& ASR $\uparrow$ & 0.12 & 0.31 & 0.46 & \textbf{0.53} \\
& PTP (kW) $\downarrow$ & 0.49 & \textbf{0.27} & 0.44 & 0.42 \\
& AMD (m) $\downarrow$ & 0.30 & \textbf{0.15} & 0.16 & \textbf{0.15} \\
\midrule
\multirow{3}{*}{$[1.50,3.00]$}
& ASR $\uparrow$ & 0.11 & 0.42 & 0.72 & \textbf{0.83} \\
& PTP (kW) $\downarrow$ & 0.45 & \textbf{0.26} & 0.33 & 0.33 \\
& AMD (m) $\downarrow$ & 0.33 & 0.19 & 0.17 & \textbf{0.16} \\
\bottomrule
\end{tabular}
\begin{tablenotes}[flushleft]
\scriptsize
\item[] ASR: avoidance success rate;
PTP: peak total power (kW);
AMD: average movement distance (m).
PTP and AMD are averaged over successful trials.
Best values are bold.
\end{tablenotes}
\end{threeparttable}
}}
\end{table}

Table~\ref{tab:3} and Fig.~\ref{fig:ablation-results} summarize the ablation results. Motion selection is essential to matching the reference response to the threat. Random selection reduces ASR to 0.02–0.12 across groups and remains substantially less successful even in the long-arrival-time group. Nominal phase progression yields ASR values of 0.08, 0.31, and 0.42. Its long-group result remains well below BARE's 0.83, supporting the benefit of learned phase adaptation over nominal phase progression. Removing locality reward yields 0.08, 0.46, and 0.72, showing that the reward improves success on top of the existing reference motions. These results support combining threat matching, execution timing adaptation, and the locality reward in BARE.

\section{Real-World Experiments}

\subsection{Experimental Setup}

We deploy BARE on Unitree G1. Onboard estimation provides proprioception, while OptiTrack tracks obstacles equipped with motion-capture markers; velocities are estimated from successive positions and transformed into the robot frame. An operator attacks from the front, sides, and rear using ball-on-rod thrusts and sweeps, kicks, disc throws, and horizontal fist sweeps.

\subsection{Real-Robot Results}
The robot starts from a stable standing posture, performs an evasive maneuver when an obstacle approaches, and then returns to stable standing. Figure \ref{fig:real-demo} presents representative responses to attacks targeting the upper, middle, and lower body regions, demonstrating reference-guided evasion and subsequent recovery on hardware. BARE achieves a 57\% avoidance success rate in real-world experiments on Unitree G1. Success follows the simulation criteria: no robot--obstacle contact during obstacle passage, no instability during evasion and recovery, and a return to stability. The performance gap between simulation and real-world experiments may stem from sensing and control latency, imperfect actuator response, and differences in ground contact dynamics, which can affect timely evasion and stable recovery. Differences between physical attacks and simulated projectiles may also contribute to this gap.

 \section{Conclusion}

Humanoid articulation offers the potential to evade incoming obstacles through local body adjustments, but coordinating these responses with whole-body stability within a short response window remains challenging. We presented BARE, a humanoid reflexive evasion framework guided by human motion priors. By selecting reference motions according to link-level collision risk and adapting their execution phase to threat urgency, BARE guides a learned policy to coordinate targeted evasion with balance and recovery. 
Simulation results demonstrate more effective evasion than the baselines across tested arrival-time groups, with less whole-body movement and lower peak power demands during successful responses. Ablations support the contributions of risk-conditioned motion selection, phase adaptation, and the locality reward. Real-world experiments on Unitree G1 demonstrate evasion and recovery against diverse threats, highlighting the potential of human motion priors and risk-guided reference adaptation for humanoid reflexive evasion.

Despite its effectiveness, our method has several limitations. The current system relies on external motion capture for obstacle states, and reliable evasion under the tightest response constraints remains challenging. Evasion on uneven terrain and under constrained support conditions also remains to be evaluated. Future work will integrate onboard sensing and active observation for autonomous threat detection and tracking, and incorporate terrain and contact states into reference adaptation to extend stable evasion to more complex environments. 

\bibliographystyle{IEEEtran}
\bibliography{references}

\end{document}